\documentclass[10pt,twocolumn,letterpaper]{article}

\usepackage{cvpr}
\usepackage{times}
\usepackage{epsfig}
\usepackage{graphicx}
\usepackage{amsmath}
\usepackage{amssymb}
\usepackage{paralist}
\usepackage{stmaryrd}
\usepackage[pagebackref=true,breaklinks=true,letterpaper=true,colorlinks,bookmarks=false]{hyperref}
\usepackage{multirow}
\usepackage{booktabs}
\usepackage{caption}

\usepackage[breaklinks=true,bookmarks=false]{hyperref}
\newcommand{\myparagraph}[1]{\vspace{-5.5pt}\paragraph{#1}}
\cvprfinalcopy 

\def\cvprPaperID{2371} 

\ifcvprfinal\fi
\begin{document}

\title{Music Gesture for Visual Sound Separation}

\author{Chuang Gan$^{1,2}$, Deng Huang$^{2}$, Hang Zhao$^{1}$, Joshua B. Tenenbaum$^{1}$, Antonio Torralba$^{1}$\\
$^1$ MIT,
$^2$ MIT-IBM Watson AI Lab}


\twocolumn[{%
\renewcommand\twocolumn[1][]{#1}%
\maketitle
    \centering
    \includegraphics[width=\textwidth]{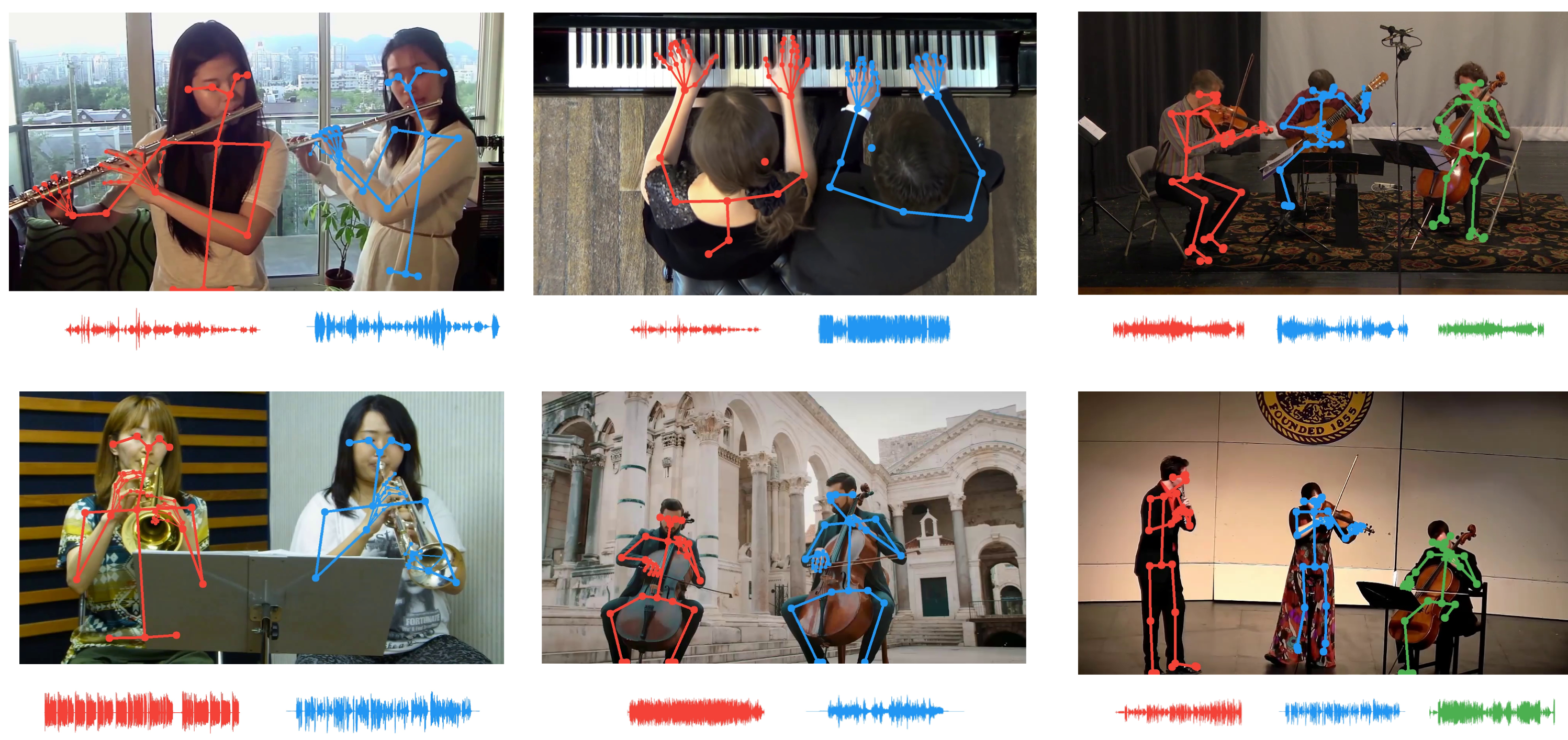}
    \captionof{figure}{We propose to leverage explicit body dynamics motion cues for visual sound separation in music performances. We show that our new model can perform well on both heterogeneous and homogeneous music separation tasks. 
    }
    \label{fig:teaser}

\vspace{5mm}
}]

\begin{abstract}
Recent deep learning approaches have achieved impressive performance on visual sound separation tasks. However, these approaches are mostly built on appearance and optical flow like motion feature representations, which exhibit limited abilities to find the correlations between audio signals and visual points, especially when separating multiple instruments of the same types, such as multiple violins in a scene. To address this, we propose ``Music Gesture," a keypoint-based structured representation to explicitly model the body and finger movements of musicians when they perform music. We first adopt a context-aware graph network to integrate visual semantic context with body dynamics, and then apply an audio-visual fusion model to associate body movements with the corresponding audio signals. Experimental results on three music performance datasets show: 1) strong improvements upon benchmark metrics for hetero-musical separation tasks (i.e. different instruments); 2) new ability for effective homo-musical separation for piano, flute, and trumpet duets, which to our best knowledge has never been achieved with alternative methods. Project page: \url{http://music-gesture.csail.mit.edu}.

\end{abstract}

\section{Introduction}

Music performance is a profoundly physical activity.  The interactions between body and the instrument in nuanced gestures produce unique sounds~\cite{godoy2010musical}. When performing, pianists may strike the keys at a lower register or ``tickle the ivory'' up high; Violin players may move vigorously through a progression while another player sways gently with a melodic base; Flautists press a combination of keys to produce a specific note. As humans, we have the remarkable ability to distinguish different sounds from one another, and associate the sound we hear with the corresponding visual perception from the musician's bodily gestures.

Inspired by this human ability, we propose ``Music Gesture" (shown in Figure~\ref{fig:teaser}), a structured keypoint-based visual representations to makes use of the body motion cues for sound source separation. Our model is built on the mix-and-separate self-supervised training procedure initially proposed by Zhao~\etal~\cite{Zhao_2018_ECCV}. Instead of purely relying on visual semantic cues~\cite{Zhao_2018_ECCV,gao2018object-sounds,gao2019co,Xu_2019_ICCV} or low-level optical-flow like motion representations~\cite{zhao2019sound}, we consider to exploit the explicit human body and hand movements in the videos. To achieve this goal, we design a new framework, which consists of a video analysis network and an audio-visual separation network. The video analysis network extracts body dynamics and semantic context of musical instruments from video frames.  The audio-visual separation network is then responsible for separating each sound source based on the visual context. In order to better leverage the body dynamic motions for sound separations, we further design a new audio-visual fusion module in the middle of the audio-visual separation network to adjust sound features conditioned on visual features.

We demonstrate the effectiveness of our model on three musical instrument datasets, URMP~\cite{li2018creating}, MUSIC~\cite{Zhao_2018_ECCV} and AtinPiano~\cite{moryossef2020at}. Experimental results show that by explicitly modeling the body dynamics through the keypoint-based structured visual representations, our approach performs favorably against state-of-the-art
methods on both hetero-musical and homo-musical separation task. In summary, our work makes the following contributions:
\begin{compactitem}
	\item We pave a new research direction on exploiting body dynamic motions with structured keypoint-based video representations to guide the sound source separation. 

	\item We propose a novel audio-visual fusion module to associate human body motion cues with the sound signals.

    \item Our system outperforms previous state-of-the-arts approaches on hetero-musical separation tasks by a large margin.
    
    \item We show that the keypoint-based structured representations open up new opportunities to solve harder homo-musical separation problem for piano, flute, and trumpet duets.
\end{compactitem}

\section{Related Work}
\label{sec:related}


\myparagraph{Sound separation.} Sound separation is a central problem in the audio signal processing area~\cite{mcdermott2009cocktail,haykin2005cocktail}, while the classic solutions for it are based on Non-negative Matrix Factorization (NMF) \cite{virtanen2007monaural,cichocki2009nonnegative,smaragdis2003non}. These are not very effective as they rely on low-level correlations in the signals.
Deep learning based methods are taking over in the recent years. Simpson~\etal~\cite{simpson2015deep} and Chandna~\etal~\cite{chandna2017monoaural} proposed CNN models to predict time-frequency masks for music source separation and enhancement.
Another challenging problem in speech separation is identity permutation: a spectrogram classification model could not deal with the case with arbitrary number of speakers talking simultaneously.
To solve this problem, Hershey~\etal~\cite{hershey2016deep} proposed Deep Clustering and Yu \etal~\cite{yu2017permutation} proposed a speaker-independent training framework. 

\myparagraph{Visual sound separation.} Our work falls into the category of visual sound separation. 
Early works~\cite{barzelay2007harmony} leveraged the tight associations between audio and visual onset signal to perform audio-visual sound attribution.
Recently, Zhao~\etal~\cite{Zhao_2018_ECCV} proposed a framework that learns from unlabeled videos to separate and localize sounds with the help of visual semantics cues.  Gao~\etal~\cite{gao2018object-sounds} combined deep networks with NMF for sound separation. Ephrat~\etal~\cite{ephrat2018looking} and Owens~\etal~\cite{owens2018audio} proposed to used vision to improve the quality of speech separation.  Xu~\etal~\cite{Xu_2019_ICCV} and Gao~\etal~\cite{gao2019co} further improved the models with recursive models and co-separation loss. Those works all demonstrated how semantic appearances could help with sound separation. However, these methods have limited capabilities to capture the motion cues, thus restricts their applicability to solve harder sound source separation problems.  

Most recently, Zhao~\etal~\cite{zhao2019sound} proposed to leverage temporal motion information to improve the vision sound separation. However, this algorithm has not yet seen wide applicability to sound separation on real mixtures. This is primarily due to the trajectory and optical flow like motion features they used are still limited to model the human-object interactions, thus can not provide strong visual conditions for the sound separation. Our work overcomes these limitations in that we study the explicit body movement cues using structured keypoint-based structured representations for audio-visual learning, which has never been explored in the audio-visual sound separation tasks.

\myparagraph{Audio-visual learning.} With the emergence of deep neural networks, bridging signals of different modalities becomes easier. A series of works have been published in the past few years on audio-visual learning. By learning audio model and image model jointly or separately by distillation, good audio/visual representations can be achieved~\cite{owens2016ambient,aytar2016soundnet,arandjelovic2017look,long2018attention,long2018multimodal,gan2015devnet,korbar2018co}. Another interesting problem is sounding object localization, where the goal is to associate sounds in the visual input spatially~\cite{izadinia2013multimodal,Hershey1999,arandjelovic2017objects,senocak2018learning,Zhao_2018_ECCV}.
Some other interesting directions include biometric matching~\cite{nagrani2018seeing}, sound generation for videos \cite{zhou2017visual}, auditory vehicle tracking~\cite{gan2019self}, emotion recognition~\cite{albanie2018emotion}, audio-visual co-segmentation~\cite{rouditchenko2019self}, audio-visual navigation~\cite{gan2019look}, and 360/stereo sound from videos~\cite{gao20182,morgado2018self}.



 
\myparagraph{Audio and body dynamics.}
\begin{figure*}[t]
   \centering
   \includegraphics[width = 1\linewidth]{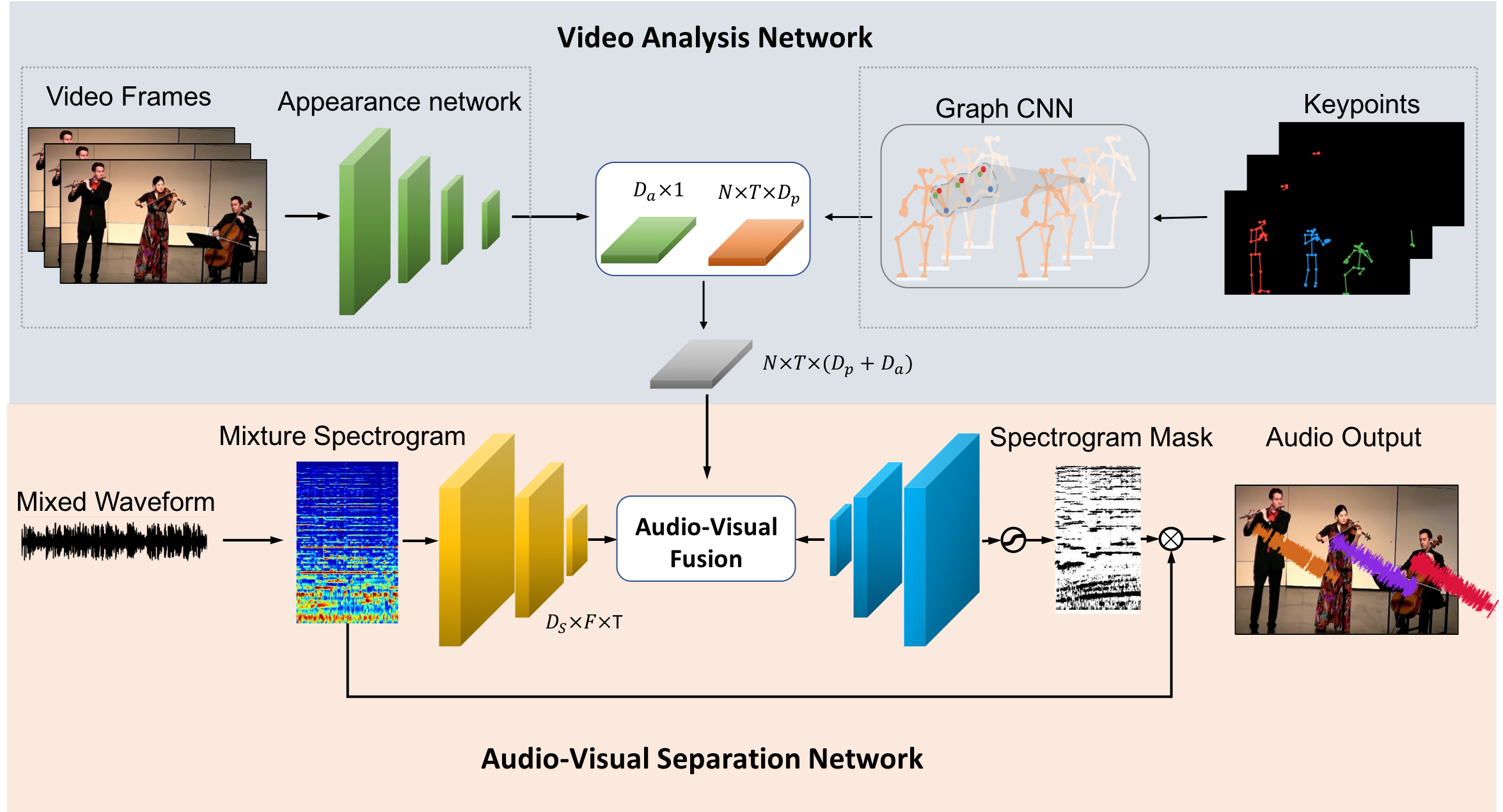}

   \caption{\textbf{An overview of our model architecture.} It consists of two components: a video analysis network and a visual-audio separation network. The video analysis network first takes video frames to extract global context and keypoint coordinates; Then a GCN is applied to integrate the body dynamic with semantic context, and outputs a latent representation. Finally, an audio-visual separation network separates sources form the mixture audio conditioned on the visual features.}
   \label{fig:framework}
\end{figure*}

There are numerous works to explore the associations between speech and facial movements~\cite{bregler1997video,brand1999voice}.  Multi-model signals extracted from face and speech has been used to do facial animations using speech~\cite{karras2017audio,taylor2017deep}, generate high-quality talking face from audio~\cite{suwajanakorn2017synthesizing,jamaludin2019you} separate mixed speech signals of multiple speakers~\cite{ephrat2018looking}, on/off screen audio source separation\cite{owens2018audio}, and lip reading from raw videos~\cite{chung2017lip}. In contrast, the correlations between body pose with sound were less explored. The most relevant to us are recent works on predicting body dynamics from music~\cite{shlizerman2018audio} and body rhythms from speech~\cite{ginosar2019learning}. This is the inverse of our goal to separate sound sources using body dynamic cues.

\section{Approach}
We first formalize the visual sound separation task and summarize our system pipeline in Section~\ref{sec: overview}. Then we present the video analysis network for learning structured representation (Section~\ref{sec: video}) and audio-visual separation model (Section~\ref{sec: audio}). Finally, we introduce our training objective and inference procedures in Section \ref{sec: training}.

\subsection{Pipeline Overview}
\label{sec: overview}
Our goal is to associate the body dynamics with the audio signals for sound source separation. We adopt the commonly used ``mix-and-separate" self-supervised training procedure introduced in~\cite{Zhao_2018_ECCV}. The main idea of this training procedure is to create synthetic training data by mixing arbitrary sound sources from different video clips. Then the learning objective is to separate each sound from mixtures conditioned on its associated visual context.

Concretely, our framework consists of two major components: a video analysis network and a audio-visual separation network (see Figure~\ref{fig:framework}). During training, we randomly select $N$ video clips with paired video frames and audio signal $\{V_k, S_k\}$, and then mix their audios by linear combinations of the audio inputs to form a synthetic mixture $S_{mix} = \sum_{k=1}^N S_k$. Given a video clip $V_k$, the video analysis network extracts global context and body dynamic features from videos. The audio-visual separation network is then responsible for separating its audio signal $S_{k}$ from the mixture audio $S_{mix}$ conditioned on the corresponding visual context $V_k$. To be noted, we trained the neural network in a supervised fashion, but it learned from unlabeled video data. Therefore, we consider the training pipeline as self-supervised learning. 

\subsection{Video Analysis Network} 
\label{sec: video}

Our proposed video analysis network integrates keypoint-based structured visual representations, together with global semantic context features. 

\myparagraph{Visual semantic and keypoint representations.} To extract global semantic features from video frames, we use ResNet-50 \cite{he2016deep} to extract the features after the last spatial average pooling layer from the first frame of each video clip. Therefore, we obtain a 2048-dimensional context feature vector for each video clip. We also aim to capture the explicit movement of the human body parts and hand fingers through the keypoint representations. To achieve that, we adopt the AlphaPose toolbox~\cite{fang2017rmpe} to estimate the 2D locations of human body joints. For estimation of hand pose, we first apply a pre-trained hand detection model and then use the OpenPose~\cite{cao2018openpose} hand API~\cite{simon2017hand} to estimate the coordinates of hand keypoints. As a result, we extract 18 keypoints for human body and 21 keypoints for each hand. Since the keypoints estimation in videos in the wild is challenging and noisy, we maintain both 2D coordinates $(X, Y)$ and the confidence score of each estimated keypoint.

\myparagraph{Context-Aware Graph CNN.}
Once the visual semantic feature and keypoints are extracted from the raw video, we adopt a context-aware Graph CNN (CT-GCN) to fuse the semantic context of instruments and human body dynamics. This architecture is designed for the non-grid data, suitable for explicitly modeling the spatial-temporal relationships among different keypoints on the body and hands.

The network architecture design is inspired by previous work on action recognition~\cite{yan2018spatial} and human shape reconstruction~\cite{kolotouros2019convolutional}. Similar to~\cite{yan2018spatial}, we start by constructing a undirected spatial-temporal graph $G = \{V, E\}$ on a human skeleton sequence. In this graph, each node $v_i \in \{V\}$ corresponds to a keypoint of the human body; edges reflect the natural connectivity of body keypoints. 

The input features for each node is represented as 2D coordinates and the confidence score of a detected keypoint over time $T$. To model the spatial-temporal body dynamics, we first apply a Graph Convolution Network to encode the pose at each time step independently. Then, we perform a standard temporal convolution on the resulting tensor to fuse the temporal information. The encoded pose feature $f_v$ is defined as follows:
\begin{equation}
 f_v = \hat{A} X W_s W_t,
\end{equation}
where $X \in R^{N \times T \times D_n}$ is the input features, $W_s$ and $W_t$ are the weight matrices of spatial graph convolution and 2D convolution, and $\hat{A} \in R^{N \times N}$ is the row-normalized adjacency matrix of the graph; $N$ represents the number of keypoints; $D_n$ represents the feature dimension for each input node. Inspired by previous work~\cite{yan2018spatial}, we define the adjacency matrix based on the joint connections of the body and fingers.  The output of the GCNs is updated features of each keypoint node. 

To further incorporate the visual semantic cues, we concatenated the visual appearance context features to each node feature as the final output of the video analysis network. The context-aware graph CNN is capable of modeling both semantic context and body dynamics, thus providing strong visual cues to guide sound separations. There could be other model designs options. We leave this to future work.

\subsection{Audio-Visual Separation Network}
\label{sec: audio}
Finally, we have an audio-visual separation network, which takes the spectrogram of mixture audio with visual representation produced by the video analysis network as input, to predict a spectrogram mask and generate the audio signal for the selected video.

\myparagraph{Audio Network.} We adopt a U-Net style architecture~\cite{ronneberger2015u}, namely an encoder-decoder network with skip connections for the audio network. It consists of 4 dilated convolution layers and 4 dilated up-convolution layers. All dilated convolutions and up-convolutions use 3 $\times$ 3 spatial filters with stride 2, dilation 1 and followed by a BatchNorm layer and a Leaky ReLU. The input of the audio network is a 2D time-frequency spectrogram of mixture sound and the output is a same-size binary spectrogram mask. We infuse the visual features into the middle part of the U-Net for guiding the sound separation.

\begin{figure}[b]
   \centering
   \includegraphics[width = 1\linewidth]{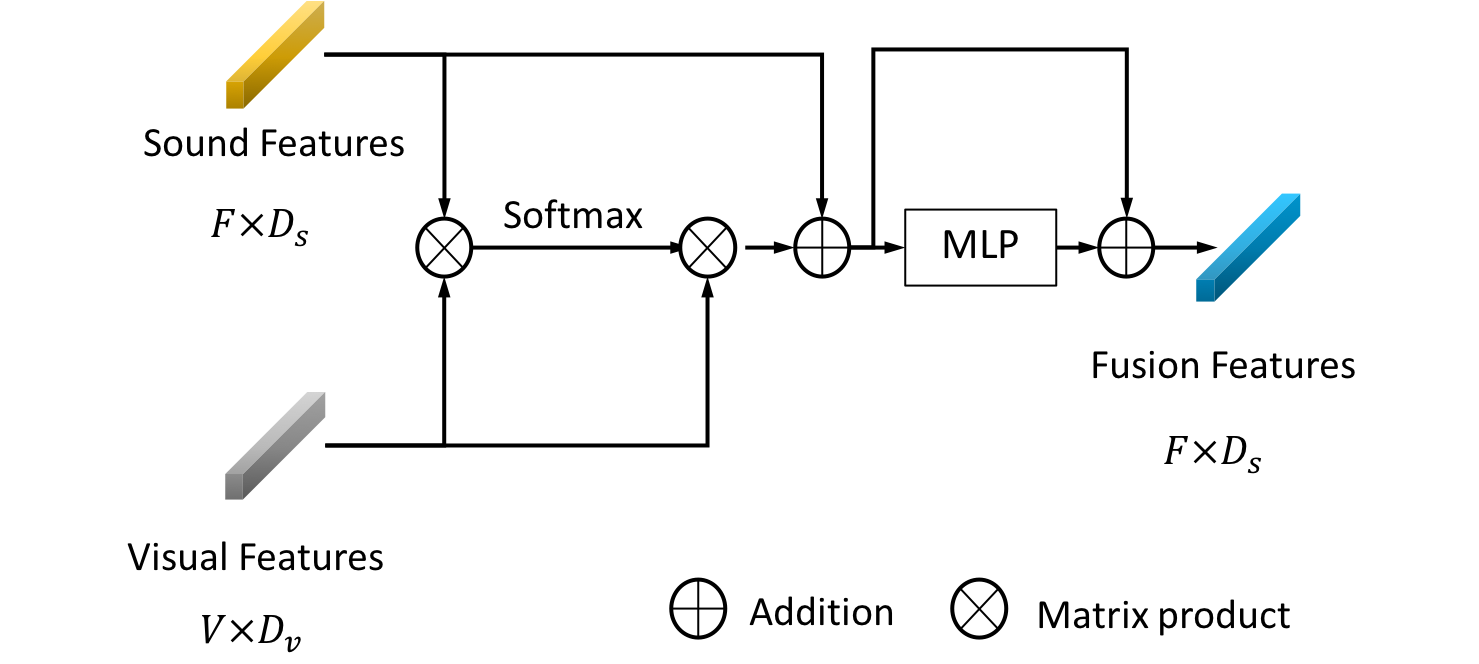}

   \caption{Audio-visual fusion module of the model in  Figure ~\ref{fig:framework}.}
   \label{fig:fusion}
\end{figure}

\myparagraph{Audio-visual fusion.}
To better leverage body dynamic cues to guide the sound separation, we adopt a self-attention~\cite{vaswani2017attention} based cross-modal early fusion module to capture the correlations between body movement with the sound signals. As shown in Figure~\ref{fig:fusion} , the fused feature $z_t$ at each time step $t$ is defined as follows:
\begin{equation}
    h_t = Softmax(f_s^t \cdot {f_v^t}^T)f_v^t + f_s^t,
\end{equation}
\begin{equation}
    z_t = MLP(h_t) + h_t,
\end{equation}
where $f_s^t \in R^{F \times D_s}$ and $f_v^t \in R^{N  \times D_v}$ represents visual and sound features at time step $t$. $F$, $D_v$, and $D_s$ denote the frequency bases of the sound spectrogram, the dimensions of visual features, and the dimension of sound features, respectively.  The softmax computation is along the dimension of visual feature channels. The visual feature is then weighted by attention matrix and concatenated with the sound feature. We further add a multi-layer perceptron (MPL) with residual connection to produce the output features. The MLP is implemented with two fully-connected layers with a ReLU activation function. This attention mechanism enforces the model to focus more on the discriminative body keypoints, and associate them with the corresponding sound components on the spectrogram. 

\subsection{Training and Inferences} 
\label{sec: training}
The learning objective of our model is to estimate a binary mask $M_k$. The ground truth mask of $k$-th video is calculated whether the target sound is the dominant component in the input mixed sound on magnitude spectrogram $S$,  \ie,
\begin{equation}
	M_k(t, f) = \llbracket S_k(t, f) \ge S_{mix}(t, f)\rrbracket, \quad \forall k=(1,...,N),
\end{equation}
where $(t, f)$ represents the time-frequency coordinates in the sound spectrogram.  The network is trained by minimizing the per-pixel sigmoid cross entropy loss between the estimated masks and the ground-truth binary masks. Then the predicted mask is thresholded and multiplied with the input complex STFT coefficients to get a predicted sound spectrogram.  Finally, we apply an inverse short-time Fourier Tranform (iSTFT) with the same transformation parameters on the predicted spectrogram to reconstruct the waveform of separated sound.

During testing, our model takes a single realistic multi-source video to perform sound source separation. We first localize human in the video frames. For each detected person, we use the video analysis network to extract visual feature to isolate the portion of the sound belonging to this musician from the mixed audio.

\section{Experiments}

In this section, we discuss our experiments, implementation details, comparisons and evaluations. 

\subsection{Dataset}
We perform experiments on three video music performance datasets, namely MUSIC-21\cite{zhao2019sound}, URMP~\cite{li2018creating} and AtinPiano~\cite{moryossef2020at}. MUSIC-21 is an untrimmed video dataset crawled by keyword query from Youtube. It contains music performances belonging to 21 categories. This dataset is relatively clean and collected for the purpose of training and evaluating visual sound source separation models.  URMP~\cite{li2018creating} is a high quality multi-instrument video dataset recorded in studio and provides ground truth labels for each sound source. AtinPiano~\cite{moryossef2020at} is a dataset where the piano video recordings are filmed in a way that camera is looking down on the keyboard and hands.

\begin{table*}[t]
	\begin{center}
	\begin{tabular}{lcccccc}
		\specialrule{.2em}{.1em}{.1em}
        \multirow{2}{*}{Methods} &  \multicolumn{2}{c}{2-Mix} & \multicolumn{2}{c}{3-Mix}  \\ 
    \cmidrule(lr){2-3}\cmidrule(lr){4-5}
        & SDR &  SIR   & SDR &  SIR   \\ 
        
        \hline
        			NMF~\cite{virtanen2007monaural} & 2.78   & 6.70    & 2.01   & 2.08    \\
			Deep Separation~\cite{chandna2017monoaural}  & 4.75 & 7.00 & -     & -   \\
			MIML~\cite{gao2018object-sounds} & 4.25 & 6.23  & -     & -    \\
			Sound of Pixels~\cite{Zhao_2018_ECCV} & 7.52 & 13.01  &3.65   &  8.77    \\ 
		    Co-Separation~\cite{gao2019co} & 7.64 &  13.8 & 3.94 & 8.93 \\
			Sound of Motion~\cite{zhao2019sound} & 8.31 & 14.82  &  4.87   & 9.48     \\ 
			
			\hline
			Our  & \textbf{10.12} & \textbf{15.81}   &  \textbf{5.41} &  \textbf{11.47} \\

		\specialrule{.1em}{.05em}{.05em}
	\end{tabular}
	\end{center}
	\caption{Sound source separation performance ($N=2, 3$ mixture) on different instruments.  Compared to previous approaches, our models with body dynamic motion information perform better in sound separation.}
	\label{tab:different}
\end{table*}

\subsection{Hetero-musical Separation}

 We first evaluate the model performance in the task of separating sounds from different kinds of instruments on the MUSIC dataset.

\myparagraph{Baseline and evaluation metrics}

We consider 5 state-of-the-art systems to compare against. 

\begin{compactitem}
\item \textbf{NMF}~\cite{virtanen2007monaural} is a well established pipeline for audio-only source separation based on matrix factorization;

\item \textbf{Deep Separation}~\cite{chandna2017monoaural} is a CNN-based audio-only source separation approach;  

\item  \textbf{MIML} \cite{gao2018object-sounds} is a model that combines NMF decomposition and multi-instance multi-label learning;

\item  \textbf{Sound of Pixels}~\cite{Zhao_2018_ECCV} is a pioneering work that uses vision for sound source separations;

\item \textbf{Co-separation}~\cite{gao2019co} devices a new model that incorporates an object-level co-separation loss into the mix-and-separate framework~\cite{Zhao_2018_ECCV};

\item  \textbf{Sound of Motions}~\cite{zhao2019sound} is a recently proposed self-supervised model which leverages trajectory motion cues.
\end{compactitem}

We adopt the blind separation metrics, including signal-to-distortion ratio (SDR), and signal-to-interference ratio (SIR) to quantitatively compare the quality of the sound separation. The results reported in this paper were obtained by using the open-source \texttt{mir\_eval}  \cite{raffel2014mir_eval} library.

\myparagraph{Experimental Setup}

Following the experiment protocol in Zhao \etal ~\cite{zhao2019sound}, we split all videos on MUSCI dataset into a training set and a test set.  We train and evaluate our model using mix-2 and mix-3 samples, which contain 2 and 3 sound sources of different instruments in mixtures. Since the real mix video data with multiple sounds on the MUSIC dataset do not have ground-truth labels for quantitative evaluation, we construct a synthetic testing set by mixing solo videos. The result of model performances are reported on a validation set with 256 pairs of sound mixtures, the same as~\cite{zhao2019sound}. We also perform a human study on the real mixtures on MUSIC and URMP dataset to measure human's perceptual quality.

\myparagraph{Implementation Details}
 
We implement our framework using Pytorch. We first extract a global context feature from a video clip using ResNet-50~\cite{he2016deep} and the coordinates of body and hand key points for each frame using OpenPose~\cite{cao2018openpose} and AlphaPose~\cite{fang2017rmpe}.
 Our GCN model consists of 11-layers with residual connections. When training the graph CNN network, we first pass the keypoint coordinates to a batch normalization layer to keep the scale of the input same. During training, we also randomly move the coordinates as data augmentation to avoid overfitting.
 
 For the audio data pre-processing, we first re-sample the audio to 11KHz. During training, we randomly take a 6-second video clip from the dataset. The audio-visual separation network takes a 6-second mixed audio clip as input, and transforms it into spectrogram by Short Time Fourier Transform (STFT). We set the frame size and hop size as  1022 and 256, respectively. The spectrogram is then fed into a U-Net with 4 dilated convolution and 4 deconvolution layers. The ouput of U-Net is an estimated binary mask. We set a threshold of 0.7 to obtain a binary mask, and then multiply it with the input mixture sound spectrogram. An iSTFT with the same parameters as the STFT is applied to obtain the final separated audio waveforms.

We train our model using SGD optimizer with 0.9 momentum. The audio separation Network and the fusion module use a learning rate of 1e-2; the ST-GCN Network and Appearance Network use a learning rate of 1e-3.

\myparagraph{Quantitative Evaluation.}

Table~\ref{tab:different} summarizes the comparison results against state-of-the-art methods on MUSIC. We observe that our method consistently outperforms all baselines in separation accuracy, as captured across metrics. Remarkably, our system outperforms a previous state-of-the-art algorithm~\cite{zhao2019sound} by 1.8dB on 2-mix and 0.6dB on 3-mix source separation in term of SDR score. These quantitative results suggest that our model can successfully exploit the explicit body dynamic motions to improve the sound separation quality.

\myparagraph{Qualitative evaluation on real mixtures.} 

Our quantitative results demonstrate that our model achieves better results than baselines.  However, these metrics are limited in their ability to reflect the actual perceptual quality of the sound separation result on real-world videos. Therefore, we further conduct a subjective human study using real mixture videos from MUSIC and URMP datasets on Amazon Mechanical Turk (AMT).  

Specifically, we compare sound separation results of our own model with best baseline system~\cite{zhao2019sound}\footnote{The results on real mixture are provided by their authors.} The AMT workers are required to compare these two systems and answer the following question: ``Which sound separation result is better?." We randomly shuffle the orders of two models to avoid shortcut solutions. Each job is performed independently by 3 AMT workers.  Results are shown in Table~\ref{tab:eval_amt_d} using majority voting. From this table, we find workers favor our system for both 2-mix and 3-mix sound separation.

\begin{table}[t]
	\begin{center}
		\begin{tabular}{c|c|c}
		\specialrule{.2em}{.1em}{.1em}
			Method & 2-Mix  & 3-Mix\\ \hline
			Sound of Motions~\cite{zhao2019sound} & 24\% &  16\% \\ \hline
			 Ours     & 76\% & 84\%  \\ 
		\specialrule{.1em}{.05em}{.05em}

			\end{tabular}
	\end{center}
	\vspace{-3mm}
	\caption{Human evaluation results for the sound source separation on mixtures of the different instruments.}
		\label{tab:eval_amt_d}

\end{table}

\subsection{Ablated study}

In this section, we perform in-depth ablation studies to evaluate the impact of each component of our model.

\myparagraph{Keypoint-based representation.} The main contribution of our paper is to use explicit body motions through keypoint-based structure representations for source separation. To further understand the ability of these representations, we conduct an ablated study using the keypoint-based structure representation only, without the RGB context features. Interestingly,we can observe that keypoint-based representations alone could also achieve very strong results (see Table ~\ref{tab:ablated}).  We hope our findings could inspire more works using structured keypoint-based representations for the audio-visual scene analysis tasks. 

\begin{table}[t]
	\begin{center}
		\begin{tabular}{l|cc}
			\specialrule{.2em}{.1em}{.1em}
			Method & SDR \\ \hline
			Ours w/o fusion   & 9.64  \\ \hline
			Ours w/o RGB & 10.22    \\ \hline
			Our  &  10.12   \\ \hline
			\specialrule{.1em}{.05em}{.05em}
		\end{tabular}
	\end{center}
	\vspace{-3mm}
	\caption{Ablated study on SDR metric for  mixtures of 2 different instruments .}
	\label{tab:ablated}
	\end{table}

\myparagraph{Visual-Audio Fusion Module.}   We  propose a novel attention based audio-visual fusion model. To verify its efficacy, we replace this module with  Feature-wise Linear Modulation (FiLM)~\cite{perez2017film} used in~\cite{zhao2019sound}. The comparison results are shown in Table~\ref{tab:ablated}.  We can find that the proposed audio-visual fusion module brings 0.5dB improvement in term of SDR metric on 2-mix sound source separation. 

\subsection{Homo-musical Separation}
In this section, we conduct experiments on a more challenging task, sound separation when sound is generated by the same instruments.

\myparagraph{Experiment Setup}

We select 5 kinds of musical instruments whose sounds are closely related to body dynamic:  trumpet, flute, piano, violin, and cello for evaluation.

Inspired by previous work~\cite{zhao2019sound,owens2018audio}, we also adopt a 2-stage curriculum learning strategy to train the sound seperation model of the same instruments. In particular, we first pre-train the model on multiple instrument separation, then learn to separate the same instrument. We compare our model against SoM~\cite{zhao2019sound}, since previous appearance based models fail to produce meaningful results in this challenging setting. The results are measured by both automatic SDR scores and human evaluations on AMT.

\myparagraph{Results Analysis.}
Results are shown in Table~\ref{tab:eval_same} and Table~\ref{tab:eval_same_amt}. From these tables, we have three key observations: 1) our proposed model consistently outperforms the SoM system~\cite{zhao2019sound} for all five instruments measured by both automatic and human evaluation metrics; 2) The quantitative results on separating violin and cello duets are close (See Table~\ref{tab:eval_same}). However, we find that the SoM system is quite brittle when testing on the real mixtures. People tend to vote our system more on real mixtures, as shown in Table~\ref{tab:eval_same_amt}; 3) The SoM provides much inferior results on trumpet, piano, and flute duets compared to our model, since the gap is larger than 3 dB. This is not very surprising since separating duet of these three instruments mainly relies on hand pose movements. It is very hard for the trajectory and optical flow features to capture such fine-grained hand movements.  Our approach can overcome this challenge in that we explicit model the body motions by tracking the coordinates changes of hand keypoints. These results further validate the efficacy of body dynamics motions on solving more and harder visual sound separation problems.  
\begin{table}[t]
	\begin{center}
		\begin{tabular}{l|ccccc}
			\specialrule{.2em}{.1em}{.1em}
			Instrument & SoM~\cite{zhao2019sound} & Ours\\ \hline
			 
		    trumpet & 1.8 & \textbf{4.9}\\
		    flute & 1.5 & \textbf{5.3}\\
		    piano & 0.8 & \textbf{3.8} \\
		    violin & 6.3  &  \textbf{6.7} \\ 
			cello  & 5.4   & \textbf{6.1}  \\
		\specialrule{.1em}{.05em}{.05em}
		\end{tabular}
	\end{center}
	 \vspace{-3mm}
	\caption{Sound source separation performance on duets of the same instruments under the SDR metric. }
		\label{tab:eval_same}

\end{table}

\begin{table}
	\begin{center}
		\begin{tabular}{l|ccccc}
			\specialrule{.2em}{.1em}{.1em}
			Instrument & SoM~\cite{zhao2019sound} & Ours\\ \hline
		
			trumpet & 18\%   &  82\%   \\ 
			flute   & 14\%  &  86\% \\ 
			piano &  30\% & 70\% \\
			violin &  26\% &  74\%   \\ 
			cello   & 28\% &  72\%   \\ 
			\specialrule{.1em}{.05em}{.05em}
		\end{tabular}
	\end{center}
	 \vspace{-3mm}
	\caption{Human evaluation result for the sound source separation on mixture of the same instruments.}
	\label{tab:eval_same_amt}
	\end{table}

\begin{figure}[b]
   \centering
   \includegraphics[width = 1.0\linewidth]{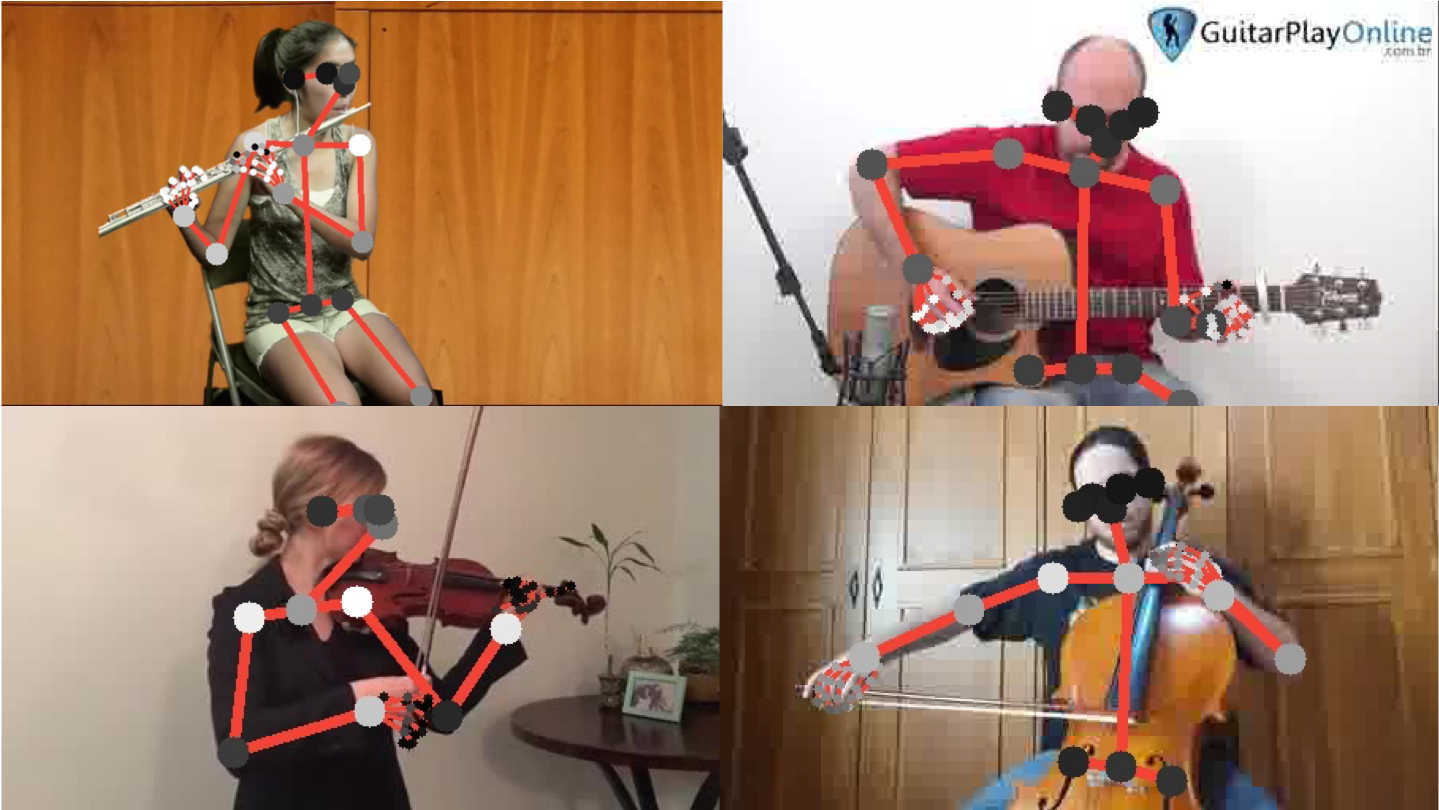}

   \caption{The attention map of body keypoints. Brighter color means higher attention score.}
   
   \label{fig:attention}
\end{figure}

\begin{figure*}[!t]
   \centering
   \includegraphics[width = 1.0\linewidth]{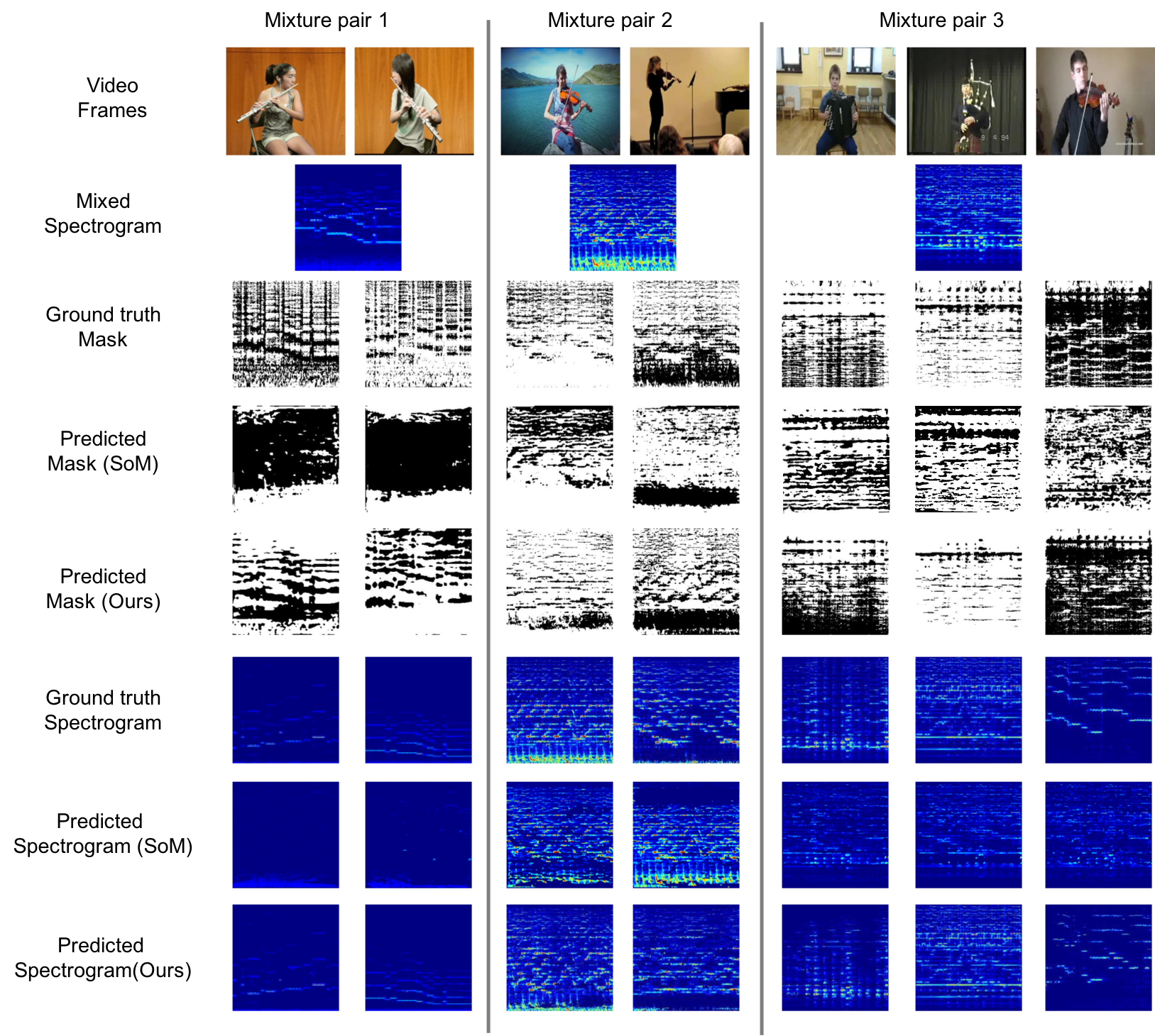}
   \vspace{-4mm}
   \caption{Qualitative results on visual sound separation compared with Sound of Motions (SoM)~\cite{zhao2019sound}.}

   \label{fig:spec}
\end{figure*}

\subsection{Visualizations}

As a further analysis, we would like to understand how body keypoints matters the sound source separation.  Figure~\ref{fig:attention} visualize the learned attention map of keypoints in the audio-visual fusion module. We observe that our model tends to focus more on hand keypoints when separating guitar and flute sounds, while pays more attention to elbows when separating the cello and violin.

Fig~\ref{fig:spec} shows qualitative results comparison between our model and the previous state-of-the-art SoM~\cite{zhao2019sound} on separating 3 different instruments and 2 same instruments.  The first row shows the video frame example, the second row shows the spectrogram of the audio mixture. The third to fifth rows show ground truth masks, masks predicted by SoM, and masks predicted by our method. The sixth to eighth rows show the ground truth spectrogram and comparisons of predicted spectrogram after applying masks on the input spectrogram. We can observe that our system produces cleaner sound separation outputs. 

Though the results are remarkable and constitute a noticeable step towards more challenging visual sound separation, our system is still far from perfect.  We observed that our method is not resilient against camera viewpoint change and body part occlusions of the musician. We conjecture that unsupervised learning of keypoints from raw images for visual sound separation might be a promising direction to explore for future work.

\section{Conclusions and Future Work}

In this paper, we show that keypoint-based structured visual representations are powerful for visual sound separation. 
Extensive evaluations show that, compared to previous appearance and low-level motion-based models, we are able to perform better on audio-visual source separation of different instruments; we can also achieve remarkable results on separating sounds of same instruments (\eg piano, flute, and trumpet), which was impossible before. We hope our work will open up avenues of using structured visual representations for audio-visual scene analysis.
In ther future, we plan to extend our approach to more general audio-visual data with more complex human-object interactions.

\myparagraph{Acknowledgement} This work is  supported by ONR MURI N00014-16-1-2007, the Center for Brain, Minds, and Machines (CBMM, NSF STC award CCF-1231216), and IBM Research.

{\small
\bibliographystyle{ieee_fullname}
\bibliography{egbib}
}

\end{document}